\documentclass[conference]{IEEEtran}
\IEEEoverridecommandlockouts
\usepackage{cite}
\usepackage{amsmath,amssymb,amsfonts}
\usepackage{algorithm}
\usepackage{algorithmic}
\usepackage{graphicx}
\usepackage{subcaption}
\usepackage{textcomp}
\usepackage{xcolor}
\usepackage{hyperref}
\usepackage{caption}
\usepackage{multirow}
\usepackage{enumerate}
\def\BibTeX{{\rm B\kern-.05em{\sc i\kern-.025em b}\kern-.08em
    T\kern-.1667em\lower.7ex\hbox{E}\kern-.125emX}}

\begin{document}

\title{Multi-Task Neural Architecture Search Using Architecture Embedding and Transfer Rank}

% \title{Conference Paper Title*\\
% {\footnotesize \textsuperscript{*}Note: Sub-titles are not captured for https://ieeexplore.ieee.org  and
% should not be used}
% \thanks{Identify applicable funding agency here. If none, delete this.}
% }

\author{\IEEEauthorblockN{Anonymous Authors}}
\author{\IEEEauthorblockN{1\textsuperscript{st} TingJie Zhang}
\IEEEauthorblockA{
\textit{School of Mathematics and Statistics} \\
\textit{Guangdong University of Technology}\\
Guangzhou, China\\
% ztj@mail2.gdut.edu.cn
}
\and
\IEEEauthorblockN{2\textsuperscript{st} HaiLin Liu}
\IEEEauthorblockA{
\textit{School of Mathematics and Statistics} \\
\textit{Guangdong University of Technology}\\
Guangzhou, China\\
% hlliu@gdut.edu.cn
}
}

\maketitle

\begin{abstract}
Multi-task neural architecture search (NAS) enables transferring architectural knowledge among different tasks. However, ranking disorder between the source task and the target task degrades the architecture performance on the downstream task. We propose KTNAS, an evolutionary cross-task NAS algorithm, to enhance transfer efficiency. Our data-agnostic method converts neural architectures into graphs and uses architecture embedding vectors for the subsequent architecture performance prediction. The concept of transfer rank, an instance-based classifier, is introduced into KTNAS to address the performance degradation issue. We verify the search efficiency on NASBench-201 and transferability to various vision tasks on Micro TransNAS-Bench-101. The scalability of our method is demonstrated on DARTs search space including CIFAR-10/100, MNIST/Fashion-MNIST, MedMNIST. Experimental results show that KTNAS outperforms peer multi-task NAS algorithms in search efficiency and downstream task performance. Ablation studies demonstrate the vital importance of transfer rank for transfer performance.
\end{abstract}

\begin{IEEEkeywords}
neural architecture search, evolutionary multi-tasking optimization, knowledge transfer.
\end{IEEEkeywords}

\section{Introduction}
Neural architecture search (NAS) has achieved significant success in image classification \cite{he2016deep,zagoruyko2016wide,huang2017densely,sandler2018mobilenetv2}, object detection \cite{girshick2015fast} and semantic segmentation \cite{ronneberger2015u}. However, NAS works well only for a specific task. When the task changes, NAS needs to learn from scratch, which is expensive in real-world applications. 

\begin{figure}[htb]
\centerline{\includegraphics[width=0.5\textwidth]{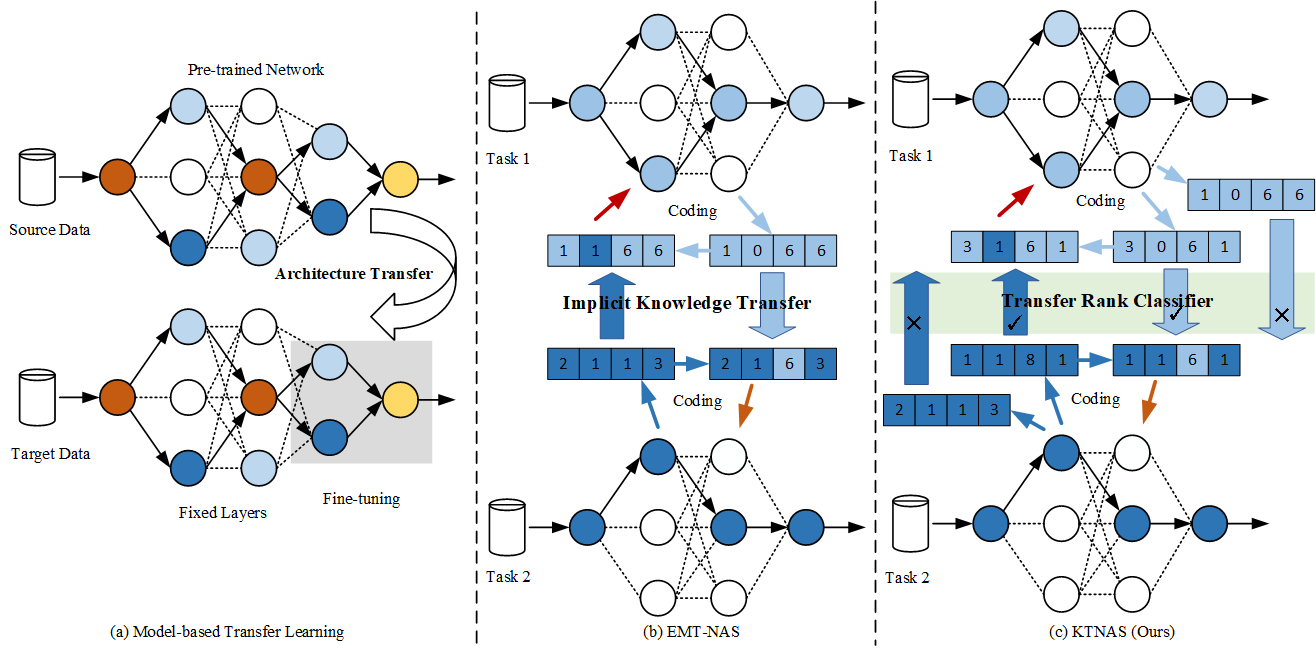}}
\caption{Different knowledge transfer processes between (a) model-based transfer learning, (b) EMT-NAS and (c) KTNAS. Best viewed in color.}
\label{kt}
\end{figure}

In such cases, knowledge transfer is applied to transfer knowledge among multiple tasks, which is desirable to reduce unnecessary search costs. Knowledge transfer involves extracting knowledge from the source tasks and applying knowledge to the target task, which is effective to enhance performance mainly due to data augmentation or model/feature reuse \cite{pan2009survey}. In Fig.~\ref{kt}a, the entire network architecture and weights acquired from the source task is reused in the target task after some steps of additional training.
%% kt

The architecture searched by NAS algorithms over different tasks exhibits similarity and transferability. For example, the backbone designed for CIFAR-10 can be easily reused by other related tasks, such as CIFAR-100 and ImageNet \cite{zoph2018learning}. In other words, architectural similarity enables the architectural knowledge extracted from the source task to be preserved, reused and refined to similar tasks \cite{zhou2023towards}. 

In addition to task-specific self-evolution, multi-task NAS further allows cross-task knowledge transfer for sharing useful architectural components over different search processes. 
As shown in Fig.~\ref{kt}b, EMT-NAS \cite{liao2023emt} utilizes the knowledge-sharing method proposed in MFEA \cite{gupta2015multifactorial} for performing crossover operation between architectures belonging to different tasks, to accelerate multiple separated search processes. Besides, the algorithm maintains a personalized architecture and corresponding weights for each task to alleviate catastrophic forgetting \cite{benyahia2019overcoming}. 
Another parallel work MTNAS \cite{liao2023emt} devises an adaptive transfer frequency to trade off the self-evolution and knowledge transfer for alleviating the potential negative transfer. 
In addition, a low-fidelity evaluation strategy is used to accelerate knowledge extraction. 

\begin{figure}[htb]
\centerline{\includegraphics[width=0.35\textwidth]{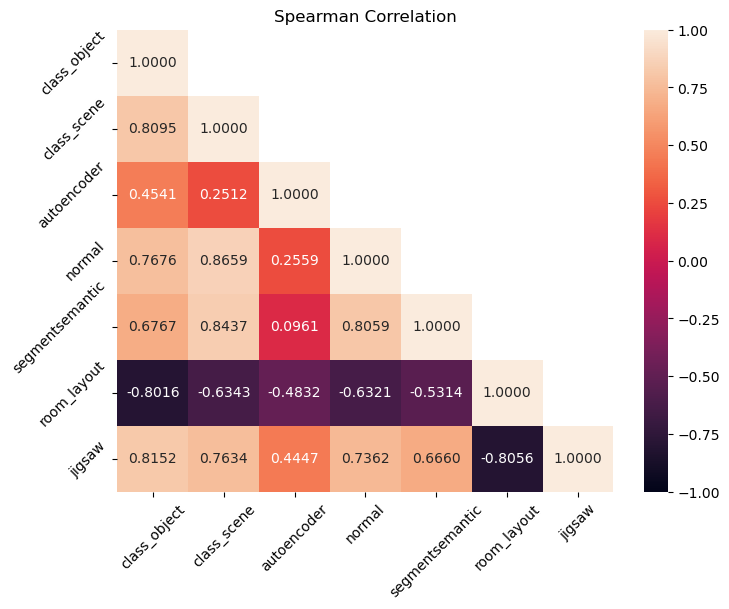}}
\caption{Ranking correlation of transferred architectures among 7 tasks on Micro TransNAS-Bench-101.}
\label{fig-rank-corr}
\end{figure}

However, ranking disorder between the source and target task weakens the architecture performance on the downstream task. As shown in Fig.~\ref{fig-rank-corr}, small ranking correlation will lead to performance degradation, namely transferred architectures perform well on the source task while poorly on the target task. Simply selecting transferred architectures by the source task ranking results in the loss of promising architectures. Not helpful transferred architectures also lead to additional computation costs or even disrupts the learning of target task. In other words, ranking disorder brings some negative influence on the learning of target task, called negative transfer. 
%% fig-rank-corr

The parts of cells that do matter for architecture performance often follow similar and simple patterns, which make them effective in transfer scenarios \cite{wan2022redundancy}. Our work seeks to identify architectures most likely to possess the common patterns, maximizing the probability of positive transfer. 

Motivated by MMOTK\cite{chen2022multiobjective} in evolutionary multi-task optimization (EMTO), we borrow the concept of transfer rank to guide the selection of transfer individuals. To reduce computational cost, we utilize the node2vec \cite{grover2016node2vec} algorithm to map a network topology into a low-dimensional feature vector for the subsequent architecture performance prediction. 

In this work, we transfer architectures with high rank to mitigate potential negative transfer. Then, we perform crossover operation between pairwise architectures to achieve cross-task knowledge sharing. The target task can reuse the actually useful architectural components to accelerate the self-evolution search process and improve the performance of learning . 

We summarize our contributions as follows:
\begin{itemize}
    \item This work introduces transfer rank into multi-task NAS for a more accurate selection of candidates and achieves effective knowledge transfer and mitigates negative transfer.
    \item To reduce computational costs, we convert architectures into graphs and utilize graph embedding vectors for the subsequent performance prediction.
    \item Extensive experiments show that our approach KTNAS outperforms its multi-task counterparts. Ablation studies and transfer performance analysis are conducted for further discussion.
\end{itemize}

We review background and related work in Section \ref{background} and present the details of proposed KTNAS in Section \ref{algo}. Experiment settings and results analysis are provided in Section \ref{expr}.

\section{Background and related work}\label{background}

\subsection{Multi-task NAS}
NAS faces various vision tasks, such as classification, scene classification, autoencoding and so on \cite{duan2021transnas}. According to search strategy, NAS can be mainly categorized into three classes: reinforcement learning (RLNAS) \cite{zoph2018learning,pham2018efficient}, gradient optimization (GONAS) \cite{liu2018darts,xie2018snas}, and evolutionary algorithm (EvoNAS) \cite{real2017large,liu2017hierarchical,real2019regularized,sun2019completely,li2021surrogate,lu2019nsga,yang2020cars,zhang2020efficient}. 
Compared with RLNAS and GONAS, EvoNAS is simple yet efficient in transferring knowledge via crossover operations.

We define transferability that effective architecture patterns contained in the source task can be transferred and reused for target task. Depending on task scenarios, knowledge transfer in NAS can be divided into the following two types. 

The first class only utilizes external knowledge to promote the target task learning. 
For instance, the cell structure learned on CIFAR-10 is transferable to other image classification tasks, such as CIFAR-100 and ImageNet \cite{zoph2018learning}. 
Similarly in Fig.~\ref{kt}a, fine-tuning, a simple type of model-based transfer learning, enables model/feature reuse. 

The second class uses bidirectional transfer of internal knowledge to improve the performance of different tasks. 
\cite{pasunuru2019continual} searches for a better generalized cell by averaging the controller awards over different tasks.
MT-ENAS \cite{cai2021multi} proposed a surrogate model based on radial basis function neural network to predict architecture performance. 
A cross-task interaction layer is devised to combine learned knowledge of multiple tasks.
EMT-NAS \cite{liao2023emt} selects the transferred solutions based on their validation accuracy on source task, which is irrational due to the ranking disorder as previously mentioned.
Another parallel work MTNAS \cite{liao2023emt} proposes adaptive transfer frequency and low-fidelity evaluation to enhance search efficiency. 

As a concurrent work, KTNAS supplements the existing knowledge-transfer strategies of multi-task NAS.
As shown in Fig.~\ref{kt}c, effective knowledge transfer in KTNAS is realized by accurate individuals selection.
Experimental results show that KTNAS can effectively alleviate negative transfer and improve the target task learning.

\subsection{EMTO}
EMTO utilizes evolutionary algorithms (EAs) to achieve knowledge transfer across different evolutions. 
In EMTO, the common practice is to build an independent population for each task \cite{gupta2015multifactorial}. 
Each population has two behaviors, task-specific self-evolution and cross-task knowledge transfer. 
For the latter, some knowledge-sharing mechanisms \cite{gupta2016multiobjective,feng2018evolutionary,lin2019multiobjective,chen2022multiobjective} are investigated that useful solutions from other tasks are identified, refined or directly injected into the target population.

When two tasks share similarity, the promising solutions belonging to one task may be helpful for another task.
MO-MFEA \cite{gupta2016multiobjective}, a multi-objective multi-factor optimization algorithm, transforms solutions into a unified space for reuse. 
EMEA \cite{feng2018evolutionary} directly transfers non-dominated solutions between highly similar tasks. 
EMTIL \cite{lin2019multiobjective} uses a Bayesian classifier with the incremental learning to divide candidates into two categories: positive-transfer and negative-transfer solutions. 
For accurate multi-level classification, MMOTK \cite{chen2022multiobjective} defines transfer rank, a supervised instance-based model, to quantify the transfer priority. 
Solutions with high transfer rank are selected for knowledge transfer.

In this work, we build a separate population for each task and introduce transfer rank to promote the effectiveness of knowledge extraction.

\subsection{Architecture embedding}
An architecture can be typically viewed as a directed acyclic graph (DAG), where the node denotes type of operations and the edge denotes connections between nodes. Graph/Architecture embedding methods map architectures with similar accuracies to the adjacent vector region.
From feature space perspective, architecture embedding reduces the feature dimension of an architecture and the resulting vectors can be used for downstream tasks. 

node2vec \cite{grover2016node2vec} devises an efficient strategy for exploring diverse neighborhoods of nodes. 
arch2vec \cite{yan2020does} performs unsupervised architecture representation learning without accuracies as labels.
CATE \cite{yan2021cate} uses Transformers with cross-attention to learn architecture encodings.

In this work, we use node2vec as architecture embedding method. Ablation study about architecture embedding selection can be found in Section \ref{param-sense}.

\section{Proposed Algorithm}\label{algo}
In this section, we first give the problem statement.
Two key components of tranfer rank, concept of positive transfer and architecture similariy representation, are discussed next.  
We give the definition, update and selection of transfer rank and finally overview the framework of KTNAS.

\subsection{Problem statement}\label{problem}
Multi-task NAS aims to optimize multiple tasks simultaneously. 
Knowledge transfer considers the optimization of architecture (corresponding to $D_{val}$). For simplicity, the training of weights (corresponding to $D_{trn}$) is neglected in the following expression.
Given the number of tasks $N$, training dataset $D_{trn}=\{D_{trn}^i\}_{i=1}^N$ and validation dataset $D_{val}=\{D_{val}^i\}_{i=1}^N$, we formulate multi-task NAS as EMTO problem:
\begin{equation}
\left\{\begin{matrix}
\alpha_1^*={argmin}_{\alpha_1}L(\omega_1(\alpha_1),\alpha_1;D^1_{val})
\\ \alpha_2^*={argmin}_{\alpha_2}L(\omega_2(\alpha_2),\alpha_2;D^2_{val})
\\ ...
\\ \alpha_N^*={argmin}_{\alpha_N}L(\omega_N(\alpha_N),\alpha_N;D^N_{val})
\\ s.t.\; \alpha_i \in \Omega, i=1,2,\ldots,N
\end{matrix}\right.
\end{equation}

where $L$ denotes task-specific loss function (cross-entropy loss for classification, mIoU for segmentation, SSIM for autoencoding), 
$\alpha^*_k,\alpha_k,\omega_k(\alpha_k)$ respectively denote the optimal architecture, the found architecture and corresponding weights for the $k$-th task, $\Omega$ denotes unified search space for all tasks.

\subsection{Concept of positive transfer}\label{positive}
When evaluating transferred architectures performance on a novel task, fully or partially training the network makes the assessment of positive transfer become extremely low efficient.
To address this problem, we utilize the strong correlation between the parent and child architectures performance to propose a positive transfer definition. 
Architectural knowledge, beneficial topology or components, will be inherited by children architecture through crossover operation. 
Therefore, we determine transferred architecture whether positive or negative transfer in an indirect way, by children architecture ranking on target task rather than its own ranking on source task.
This method avoids costly evaluations of transferred architectures, making knowledge transfer with almost no additional cost. 

\begin{figure}[t]
\centerline{\includegraphics[width=0.5\textwidth]{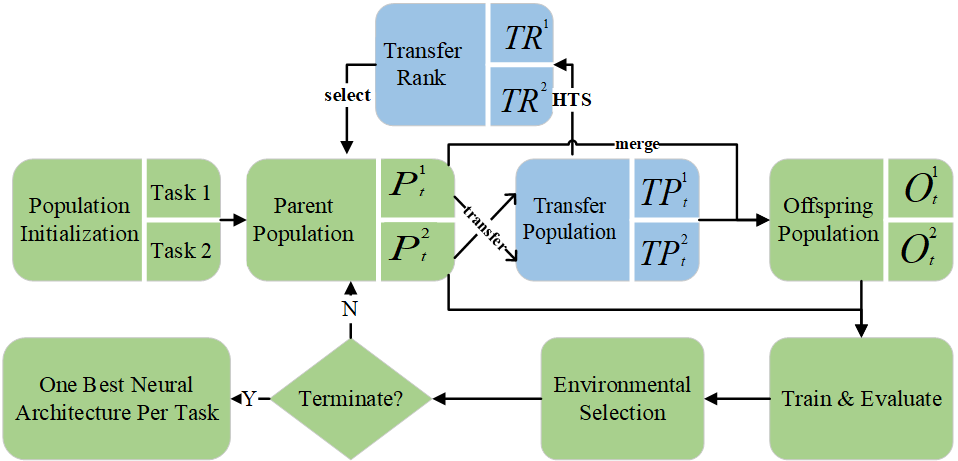}}
\caption{The flow chart of KTNAS for two-task scenarios. Green denotes self-evolution. Blue denotes cross-task knowledge transfer. Best viewed in color.}
\label{flowchart}
\end{figure}

For each task in Fig.~\ref{flowchart}, we build a parent population $P$ with population size $K$, a transfer population $TP$ extracted from other tasks. 
Offspring population $O$ is generated by applying reproduction operators on individuals of $P\cup TP$. 
Ranking ratio $r\%$, a predefined hyper-parameter, represents the ranking threshold for positive transfer assessment.
Denote the next parent generation $Z=topK(P\cup TP \cup O)$.
We define an architecture is positive transfer when its children architecture ranks in top $r\%$ of $Z$, otherwise negative transfer. 
% flowchart

\subsection{Architectural similarity representation}\label{distance}
When applying EMTO to NAS for knowledge transfer, the form of solutions comes to architectures, which is very different with traditional EMTO.
The high dimension of feature space leads to unaffordable computational costs when calculating architectural similarity.

To address this challenge, we use a fast and low-cost algorithm node2vec \cite{grover2016node2vec} to obtain architecture embedding vectors and reduce feature dimension.
Specially, an architecture is mapped into a 256-dimension feature vector, where architectural similarity is transformed into that of embedding vectors. 
To get a distance-like representation, we give the cosine distance metric $dist$ for architectural similarity as follows:
\begin{equation}
    dist(\alpha_1,\alpha_2)=1-\cos(node2vec(\alpha_1),node2vec(\alpha_2))
\end{equation}
where $\alpha_1,\alpha_2$ denote pairwise architectures, $cos$ denotes cosine similarity.
Notably, the value range of $dist$ is from 0 to 2, where a small value indicates high similarity, while a large value indicates low similarity.

\subsection{Transfer rank}\label{TR}
We focus on the application of transfer rank on knowledge extraction, namely selecting promising architectures from other tasks to transfer.
Two key components of transfer rank, positive transfer definition (described in Section \ref{positive}) and similarity metric (described in Section \ref{distance}) tailored for architecture, are discussed above.

Transfer rank aims to guide the selection of transferred individuals from other tasks to the target task. 
The core idea of transfer rank is to quantify the transfer priority by several nearest neighbors and make a more accurate classification for candidate architectures.

Algorithm \ref{alg:hts} maintains historical transferred set (HTS) for each task.
HTS records the information whether a previous transferred individual is positive or negative transfer, corresponding to the positive or negative class set.
We label positive-transfer individuals with value 1 while negative-transfer individuals with value -1.
The number of saved generations $m$, a predefined hyper-parameter, determines how many previous generations of negative-transfer individuals can be saved to the negative-class set at most. 
A larger value of $m$ means more negative samples used for classification.
% alg-HTS
\begin{algorithm}[t]
    \caption{Update HTS}
    \label{alg:hts}
    \renewcommand{\algorithmicrequire}{\textbf{Input:}}
    \renewcommand{\algorithmicensure}{\textbf{Output:}}
    \begin{algorithmic}[1]
        \REQUIRE The current generation $t$, the transfer population $TP$
        \ENSURE The historical transferred set $HTS$

        \FOR{$c$ \text{in} ${TP}$}
            \IF {c is positive transfer}
                \STATE Put $c$ in the positive-class set $Pos$
                \STATE $label(c)=1$
            \ELSE
                \STATE Put $c$ in the negative-class set ${Neg}_t$
                \STATE $label(c)=-1$
            \ENDIF
        \ENDFOR
        \IF{$t\leq m$}
            \STATE ${HTS}\leftarrow {Pos}\cup({\cup}_{h=1}^t {Neg}_h)$
        \ELSE
            \STATE ${HTS}\leftarrow {Pos}\cup({\cup}_{h=1}^m Neg_{t-m+h})$
        \ENDIF
        \STATE Output the historical transferred set of each task $HTS$
    \end{algorithmic}
\end{algorithm}
% alg-TR
\begin{algorithm}[b]
    \caption{Calculate transfer rank}
    \label{alg:TR}
    \renewcommand{\algorithmicrequire}{\textbf{Input:}}
    \renewcommand{\algorithmicensure}{\textbf{Output:}}
    \begin{algorithmic}[1]
        \REQUIRE Historical transferred set of each task $HTS=\{s_1,\ldots,s_L\}$, the current population of each task $P=\{p_1,\ldots,p_K\}$, architectural distance metric $dist$
        \ENSURE The transfer rank $\phi_j$ of each individual $p_j$ in $P$
        \STATE Calculate the similarity matrix $D_{L\times K}=dist(s_i,p_j),i=1,\ldots,L,j=1,\ldots,K$
        \STATE Set the associated subset $\Phi_j=\emptyset$ and $\phi_j=0,j=1,\ldots,K$
        \FOR{$i\leftarrow 1$ \text{to} $L$}
            \STATE $k=\mathop{\arg\min}\limits_{j} D_{i,j}$
            \STATE ${\Phi}_k={\Phi}_k\cup \{s_i\}$
        \ENDFOR
        \STATE ${\phi}_j={\sum}_{s\in{\Phi}_j}label(s),j=1,\ldots,K$
        \STATE Output the transfer rank $\phi_j$ of each individual in $P$
    \end{algorithmic}
\end{algorithm}

Algorithm \ref{alg:TR} illustrates the calculation of transfer rank.
Transfer rank is a surrogate model to predict architecture performance by utilizing its nearest neighbors stored in HTS.
With no need of training on real datasets, transfer rank exhibits low overhead and high speed.
Using our devised architectural distance metric $dist$, we first calculate the similarity matrix between the previously transferred and current individuals.
Then, we associate the two most similar individuals and update the transfer rank of candidates.

Algorithm \ref{alg:selection} presents the selection of transferred architectures.
The number of transferred individuals $M$ is also a hyper-paramter.
Parameter sensitivity analysis about $r\%, m, M$ can be found in Section \ref{param-sense}.
% alg-select
\begin{algorithm}[t]
    \caption{Select transferred individuals}
    \label{alg:selection}
    \renewcommand{\algorithmicrequire}{\textbf{Input:}}
    \renewcommand{\algorithmicensure}{\textbf{Output:}}
    \begin{algorithmic}[1]
        \REQUIRE Historical transferred set of each task $HTS$, the current population of each task $P$, transfer rank $\phi=\{{\phi}_j\}_{j=1}^K$ of $P$
        \ENSURE The transfer population $TP$
        \STATE Divide $P$ into $A_1,\ldots,A_H$ based on descending $\phi$
        \STATE Find maximum $h$ that satisfies $\left|A_1\cup\ldots\cup A_h \right|\leq M$
        \STATE $TP\leftarrow A_1\cup\ldots\cup A_h$
        
        \IF{$\left|A_1\cup\ldots\cup\ A_h\right|<M$ and $\left|A_1\cup\ldots\cup\ A_{h+1}\right|>M$}
            \STATE $B\leftarrow M-\left|A_1\cup\ldots\cup\ A_h\right|$ individuals are randomly selected from $A_{h+1}$
            \STATE ${TP}\leftarrow {TP}\cup B$
        \ENDIF
        \STATE Output the transfer population $TP$
    \end{algorithmic}
\end{algorithm}

\begin{figure}[htb]
\centerline{\includegraphics[width=0.3\textwidth]{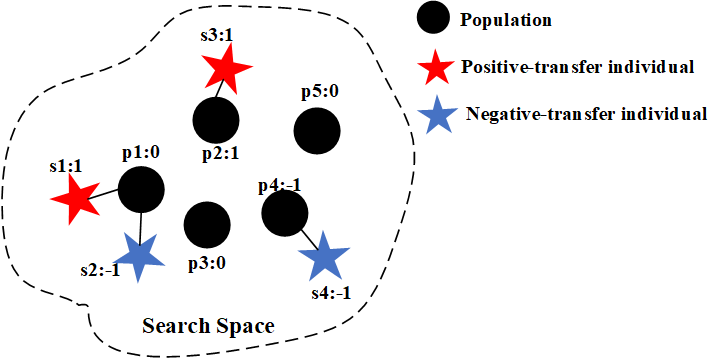}}
\caption{An example for transfer rank calculation.}
\label{fig:TR}
\end{figure}

A specific example of transfer rank is given in Fig. \ref{fig:TR}. Given current population $P=\{p_1,p_2,p_3,p_4,p_5\}$, transfer population $TP=\{s_1,s_2,s_3,s_4\}$.
\begin{enumerate}
    \item Classify the previously transferred individuals by Algorithm \ref{alg:hts}. $s_1,s_3$ are positive-transfer individuals, $s_2,s_4$ are negative-transfer individuals. 
    \item Each transferred individual in $TP$ finds the nearest individual in $P$ to associate with. For example, $s_1,s_2$ both find the nearest candidate $p_1$, $s_3$ finds $p_2$, $s_4$ finds $p_4$. 
    \item Each candidate individual in $P$ accumulates the label values of associated individuals in $TP$. For instance, $p_1$ accumulates the label value of $s_1,s_2$ as $-1+1=0$. $p_2,p_4$ directly inherits the label value of $s_3,s_4$, respectively. $p_3,p_5$ without associated individuals keep a label value of 0.
    \item Select new transferred individuals by Algorithm \ref{alg:selection}. If $M=2$, we select $p2$ and a random one from $p1,p3,p5$ to overwrite $TP$.
\end{enumerate}
% fig-TR

\subsection{KTNAS}
KTNAS aims to search for multiple tasks simultaneously. 
The flow chart of KTNAS is shown in Fig.~\ref{flowchart}. 
KTNAS is composed of random population initialization, transfer population update based on transfer rank (introduced in Section \ref{TR}), offspring population generation based on tournament selection and crossover and mutation, environmental selection by topK selection. 
% alg:KTNAS
\begin{algorithm}[t]
    \caption{The pseudo code of KTNAS}
    \label{alg:KTNAS}
    \renewcommand{\algorithmicrequire}{\textbf{Input:}}
    \renewcommand{\algorithmicensure}{\textbf{Output:}}
    \begin{algorithmic}[1]
        \REQUIRE Search space $\mathcal{A}$, an EvoNAS
        \ENSURE The best network architecture for each task

        \FOR{$i \leftarrow 1$ \text{to} $N$}
            \STATE $P_0^i\leftarrow$Randomly generate an initial population with $K$
            \STATE ${TP}_0^i\leftarrow$Generate an initial transfer population with M by randomly selecting individuals from $P_0^j(j\neq i)$
            \STATE $R_0^i\leftarrow$Train individuals of $P_0^i\cup {TP}_0^i$ on $D_{trn}^i$
            \STATE Evaluate the fitness of trained individuals in $R_0^i$ on $D_{val}^i$
            \STATE $P_1^i\leftarrow$Select top $K$ individuals of every task from $R_0^i$
            \STATE $t=1$
            \WHILE{$t<T$}
                \STATE Obtain historical transferred set ${HTS}^i_t$ by Algorithm 1
                \STATE Calculate the transfer rank of $P^i_t$ by Algorithm 2
                \STATE Generate transfer population ${TP}_t^i$ from $P_t^j(j\neq i)$ by Algorithm 3
                \STATE $O_t^i\leftarrow$Apply crossover and mutation operators on individuals of $P_t^i\cup {TP}_t^i$
                \STATE $P_t^i,O_t^i,{TP}_t^i\leftarrow$Train individuals of $P_t^i\cup O_t^i\cup {TP}_t^i$ on $D_{trn}^i$
                \STATE Evaluate the fitness of trained individuals in $O_t^i$ on $D_{val}^i$
                \STATE $P_{t+1}^i\leftarrow$Select top K individuals of every task from $P_t^i\cup O_t^i\cup {TP}_t^i$
                \STATE ${(P_{best})}_t\leftarrow$In $P_{t+1}=P_{t+1}^1\cup \ldots \cup P_{t+1}^N$, the individuals with the highest fitness in each task are evaluated on the validation dataset for the corresponding task
                \STATE $t=t+1$
            \ENDWHILE
        \ENDFOR
        \STATE Output the best individuals in $P_{best}$ of each task and decode them into the corresponding network architecture
    \end{algorithmic}
\end{algorithm}

The pseudo code of KTNAS is presented in Algorithm \ref{alg:KTNAS}. 
Recall these hyper-parameters: $N,D_{trn},D_{val},K$ in Section \ref{problem}, $r\%$ in Section \ref{positive}, $m,M$ in Section \ref{TR}.
$T$ denotes the maximum number of generations.
For each task, the initialization, training and evaluation of current population $P$ and transfer population $TP$ are performed at the zero-th generation (Line 2-6). 
The steps of Line 9-17 are repeated $T-1$ times and finally the best task-specific individuals are output (Line 20). 
First, the algorithm maintains a historical transferred set $HTS$ for each task, which saves the individual information (i.e., architecture encoding and binary transfer label) of the previous $m$ generations transferred to the current task (Line 9). 
Next, the selection of candidate individuals is conducted by transfer rank from other tasks (Line 10-11). 
Then, the crossover, mutation, training, evaluation and environmental selection in \cite{liao2023emt} are performed sequentially (Line 12-15). 
At last, the individual with the highest fitness in each task is recorded as a candidate for optimal architecture (Line 16).

\section{Experimental Results}\label{expr}
\subsection{Search spaces and hyper-parameters}
\textbf{NASBench-201 (NAS201)} \cite{dong2020bench}. NAS201 is a test suite for NAS algorithms with 15,625 candidates, where cells are generated by 4 nodes and 5 associated operation options. Three similar datasets are included, such as CIFAR-10, CIFAR-100 and ImageNet-16-120. 

\textbf{TransNAS-Bench-101 (Trans101)} \cite{duan2021transnas}. Trans101 is a benchmark dataset containing seven different vision tasks with 4096 backbones in cell-level search space. We test various NAS algorithms on Trans101 for cross-task search efficiency and generalizability.

\textbf{DARTs} \cite{liu2018darts}.
DARTs search space shows high transferability to image classification tasks. 
Following EMT-NAS\cite{liao2023emt}, we define a cell with 5 blocks and 9 optional operations, an unified search space to achieve knowledge transfer over different tasks.
We adopt the same architecture encoding and hyper-parameter settings in search and evaluation stages but different knowledge-extraction strategy (i.e., transfer rank). 
Three pairs of datasets include CIFAR-10/100, MNIST/Fashion-MNIST, MedMNIST (4 subdatasets: PathMNIST, OrganMNIST\textunderscore\{Axial,Coronal,Sagittal\}), abbreviated as C-10/100, MNIST/F-MNIST, Path, Organ\textunderscore\{A,C,S\} respectively.

\textbf{Metrics.} The first metric, the number of architecture evaluations during search stage, is used to represent search efficiency in NAS201 and Trans101. The second metric for search efficiency, "GPU Days", denotes the running time of search stage on a single GPU in DARTs.

\textbf{Experimental setup.} 
We run KTNAS 10 times with different random seeds on NAS201 and Trans101.
The mean and standard deviation of 5 independent runs are reported on DARTs, where architectures are trained from scratch.
All experiments are run on a single NVIDIA RTX 3090 24G. 
Hyper-parameters are shown in Table \ref{tab:hyper-param}. 
% tab:hyper-param
\begin{table}[h]
\captionsetup{justification=centering}
\caption{Hyper-parameters setting on 3 search spaces.}
\label{tab:hyper-param}
\begin{center}
\begin{tabular}{|c|c|c|c|}
\hline
Parameters & NAS201 & Trans101 & DARTs \\
\hline
EvoNAS method & REA \cite{real2019regularized} & GCN \cite{wen2020neural} & GA \\
Low-fidelity evaluation & zero-cost \cite{abdelfattah2021zero} & GCN-based  & weight sharing\\
Population size & 10 & 10 & 20 \\
Tournament size & 5 & 5 & 2 \\
Crossover probability & - & - & 1 \\
Mutation probability & 1 & 1 & 0.02 \\
\hline
Ranking ratio & 20\% & 20\% & 20\% \\
\# Saved generations & 5 & 5 & 5 \\
\# Transferred individuals & 4 & 4 & 4 \\
\hline
\end{tabular}
\end{center}
\end{table}

\subsection{NASBench-201 result}
We choose popular single-task NAS algorithms for comparison, such as random search (RS), regular evolution (REA) \cite{real2019regularized}, GCN \cite{wen2020neural}, LaNAS \cite{wang2021sample}, WeakNAS \cite{wu2021stronger}, Zero \cite{abdelfattah2021zero}. MTNAS \cite{zhou2023towards} is also included as baseline multi-task NAS algorithm. The optimal validation accuracy for three classification tasks (i.e., C-10, C-100 and ImageNet) is 91.61\%, 73.49\% and 46.77\%. Table \ref{tab:nas201} reports the average evaluations during search stage over 10 independent runs.
% tab:nas201
\begin{table}[h]
\captionsetup{justification=centering}
\caption{Comparison on the number of evaluations for finding the best architectures on NAS201.}
\label{tab:nas201}
\begin{center}
\begin{tabular}{|c|c|c|c|c|}
\hline
Method & C-10 & C-100 & ImageNet & Total \\
\hline
RS & 7782.1 & 7621.2 & 7726.1 & 23129.4 \\
REA \cite{real2019regularized} & 563.2 & 438.2 & 715.1 & 1439.6 \\
GCN \cite{wen2020neural} & 214.4 & 260.5 & 250.6 & 725.5 \\
LaNAS \cite{wang2021sample} & 247.1 & 187.5 & 292.4 & 727.0 \\
WeakNAS \cite{wu2021stronger} & 182.1 & 78.4 & 268.4 & 528.9 \\
Zero \cite{abdelfattah2021zero} & 230.8 & 18.4 & 213.0 & 462.2 \\
\hline
MTNAS \cite{zhou2023towards} & 106.0 & 30.1 & 92.3 & 228.4 \\
KTNAS w/o transfer rank & 105.8 & 51.6 & 66.1 & 223.5 \\
KTNAS & \textbf{100.7} & \textbf{27.9} & \textbf{60.0} & \textbf{188.5} \\
\hline
\end{tabular}
\end{center}
\end{table}

Compared with these single-task methods, KTNAS has the fewest total evaluations with 2-120$\times$. Specially, KTNAS has 120$\times$ fewer total evaluations than RS. Compared with multi-task MTNAS, KTNAS only has less search costs on each task and total evaluations. Compared to KTNAS w/o transfer rank, KTNAS has the comparable time cost on C-10 and ImageNet but has 2$\times$ fewer evaluations on C-100.

\subsection{TransNAS-Bench-101 result}
For each search process, the computational budget is set to 50 evaluations. In Table \ref{tab:trans101}, KTNAS is applied for 3 distinct vision tasks (i.e., object classification, scene classification and room layout). We report the average performance of found architectures. 
% tab:trans101
\begin{table}[h]
\captionsetup{justification=centering}
\caption{Comparison on architecture performance when limiting the number of evaluations on Trans101.}
\label{tab:trans101}
\begin{center}
\begin{tabular}{|c|c|c|c|c|}
\hline
Search & Object C. & Scene C. & Room Lay. & Average \\
Method & Acc (\%) $\uparrow$ & Acc (\%) $\uparrow$ & L2 Loss$\downarrow$ & Rank$\downarrow$ \\
\hline
RS & 45.16 & 54.41 & 61.48 & 57.7 \\
REA \cite{real2019regularized} & 45.39 & 54.62 & 61.75 & 44.0 \\
NSGA-NetV2 \cite{lu2020nsganetv2} & 45.61 & 54.75 & 61.73 & 36.3 \\
WeakNAS \cite{wu2021stronger} & 45.60 & 54.72 & 60.31 & 15.0 \\
\hline
MTNAS \cite{zhou2023towards} & 46.05 & 54.85 & 60.07 & 5.7 \\
KTNAS (ours) & \textbf{46.07} & \textbf{54.90} & \textbf{60.10} & \textbf{5.5} \\
\hline
Optimal & 46.32 & 54.94 & 59.38 & 1.0 \\
\hline
\end{tabular}
\end{center}
\end{table}

KTNAS achieve competitive performance and the highest average architecture rank than other NAS methods over 3 tasks. 
Compared with MTNAS, KTNAS can achieve comparable evaluation results on room layout (60.10\% vs. 60.07\%), while beats it on objective classification (46.07\% vs. 46.05\%) and scene classification (54.90\% vs. 54.85\%).

\subsection{DARTs results}
To demonstrate the scalability, we perform KTNAS not only on two training-free popular benchmark datasets (i.e., NAS201 and Trans101) but also on real-world DARTs search space.

\textbf{CIFAR-10/100 result.}
Multi-task algorithms search on C-10 and C-100 simultaneously, where cross-task knowledge transfer is realized by distinct EMTO methods.
Comparison results on C-10 and C-100 are in Table \ref{tab:cifar}. 
% tab-cifar
\begin{table*}[h]
\captionsetup{justification=centering}
\caption{Comparison results on C-10/100. * indicates that found architecture on C-10 is transferred to C-100, so GPU Days searched on C-10 and C-100 are the same.}
\label{tab:cifar}
\begin{center}
\begin{tabular}{|c|c|c|c|c|c|c|}
\hline
\multirow{2}{*}{Architecture} & \multirow{2}{*}{Search Method} & \multirow{2}{*}{GPU Days} & \multicolumn{2}{|c|}{C-10} & \multicolumn{2}{|c|}{C-100} \\
\cline{4-7}
 &  &  & Params (M) & Test Acc (\%) & Params (M) & Test Acc (\%)\\
\hline
Wide ResNet \cite{zagoruyko2016wide} & manual & - & 36.5 & 95.83 & * & 79.50 \\
DenseNet \cite{huang2017densely} & manual & - & 27.2 & 96.26 & * & 80.75 \\
MobileNetV2 \cite{sandler2018mobilenetv2} & manual & - & 2.1 & 94.56 & * & 77.09 \\
\hline
PNAS \cite{liu2018progressive} & SMBO & 150 & 3.2 & 96.59±0.09 & * & 80.47 \\
NASNet-A \cite{zoph2018learning} & RL & 2000 & 3.3 & 97.35 & - & - \\
ENAS \cite{pham2018efficient} & RL & 0.45 & 4.6 & 97.11 & * & 80.57 \\
DARTS \cite{liu2018darts} & GO & 1.5 & 3.3 & 97.00±0.14 & * & 79.48±0.31 \\
SNAS \cite{xie2018snas} & GO & 1.5 & 2.9 & 97.02 & - & - \\
large-scale Evo \cite{real2017large} & EA & 2750 & 5.4 & 94.6 & 40.4 & 77 \\
Hier-EA \cite{liu2017hierarchical} & EA & 300 & 64 & 96.25±0.12 & - & - \\
Amoebanet-A \cite{real2019regularized} & EA & 3150 & 3.2 & 96.66±0.06 & * & 81.07 \\
NSGA-Net \cite{lu2019nsga} & EA & 8 & 3.3 & 96.15 & * & 79.26 \\
CARS-E \cite{yang2020cars} & EA & 0.4 & 3.0 & 97.14 & - & - \\
\hline
Single-Tasking (Baseline) \cite{liao2023emt} & EA & 0.46 & 2.41 & 96.64±0.18 & 2.19 & 80.82±0.71 \\
EMT-NAS \cite{liao2023emt} & EMTO & \textbf{0.42} & 2.91 & 97.04±0.04 & 2.97 & 82.60±0.38 \\
MTNAS \cite{zhou2023towards}& EMTO & 0.50 & - & \textbf{97.25±0.04} & - & 82.55±0.45 \\
KTNAS w/o transfer rank & EMTO & 0.62 & 3.00 & 96.63±0.16 & 3.07 & 80.79±0.10 \\
KTNAS (Ours) & EMTO & 0.48 & 2.97 & 97.12±0.10 & 2.92 & \textbf{82.93±0.19} \\
\hline
\end{tabular}
\label{tab1}
\end{center}
\end{table*}

Compared with single-task methods, EMTO-based algorithms (i.e., EMT-NAS, MTNAS and KTNAS) both have higher classification accuracy, demonstrating the effectiveness of architectural knowledge transfer. 
KTNAS achieves the best performance on C-10 except for NASNet-A, but with 4166$\times$ less search costs than the latter.
KTNAS reaches comparative classification accuracy and search time, yet with a model size merely two-thirds that of ENAS. 

Compared with multi-task counterparts, KTNAS outperforms EMT-NAS by 0.33\% and MTNAS by 0.38\% in terms of C-100 test accuracy with slightly small model size.
We perform ablation study (KTNAS using random architecture selection).
Compared with KTNAS, we observe that transfer rank search for better performance and more compact model, leading to a better performance-complexity trade-off.

\textbf{MNIST/F-MNIST result.}
Comparison results are shown in Table \ref{tab:mnist}. 
The column GPU Days (\%) represents search cost relative to Single-Tasking baseline. 
Compared with manually designed models, 
EMTO-based models are competitive to hand-designed ones on MNIST, but outperform the latter on F-MNIST. 
We speculate the reason that hand-designed models are performance-saturated on simple MNIST while multi-task NAS can explore search space more effectively on difficult F-MNIST. 
% tab-mnist
\begin{table*}[h]
\captionsetup{justification=centering}
\caption{Comparison results on MNIST/F-MNIST.}
\begin{center}
\begin{tabular}{|c|c|c|c|c|c|}
\hline
\# Tasks & Model & Search Method & GPU Days (\%)$\downarrow$ & MNIST (\%) & F-MNIST (\%) \\
\hline
\multirow{5}{*}{1} & CTM \cite{granmo2019convolutional} & manual & - & 99.4 & 91.4 \\
 & FastSNN \cite{taylor2022robust} & manual & - & 99.30 & 90.57 \\
 & NeuPDE \cite{sun2020neupde} & manual& - & 99.49 & 92.4 \\
 & ResNet-18 \cite{he2016deep} & manual & - & 99.69 & 93.35 \\
 & WaveMix-128/7 \cite{jeevan2022wavemix} & manual & - & 99.71 & 93.91 \\
\hline
\multirow{4}{*}{4} & Single-Tasking (Baseline) \cite{liao2023emt} & EA & +00.00 & 99.40±0.13 & 93.70±0.42 \\
 & EMT-NAS \cite{liao2023emt} & EMTO & +2.65 & 99.40±0.09 & 93.86±0.68 \\
 & KTNAS w/o transfer rank & EMTO & +4.47 & 99.56±0.07 & 93.30±0.52 \\
 & KTNAS (Ours) & EMTO & \textbf{-10.28} & \textbf{99.62±0.17} & \textbf{94.36±0.46} \\
\hline
\end{tabular}
\label{tab:mnist}
\end{center}
\end{table*}

Compared with EMT-NAS, KTNAS acquires an enhanced accuracy of 0.22\% on MNIST and 0.50\% on F-MNIST. Notably, KTNAS achieved comparable performance with 12.93\% less GPU Days of EMT-NAS.
Ablation study also demonstrates the effectiveness of transfer rank in search efficiency and task performance. 

\textbf{Multi-tasking result on MedMNIST.}
We also study KTNAS on MedMNIST subdatasets when the number of tasks rises to 4.
Comparison results on Path, Organ\textunderscore\{A,C,S\} are shown in Table \ref{tab:med}. 
EMTO-based algorithms generally have fewer search cost than single-Tasking, proving that knowledge transfer can accelerate search stage. 
% tab-med
\begin{table*}[h]
\captionsetup{justification=centering}
\caption{Comparison results on MedMNIST. * denotes self-implementation.}
\begin{center}
\begin{tabular}{|c|c|c|c|c|c|c|c|}
\hline
\# Tasks & Model & Search Method & GPU Days (\%)$\downarrow$ & Path (\%) & Organ\textunderscore A (\%) & Organ\textunderscore C (\%) & Organ\textunderscore S (\%)\\
\hline
\multirow{6}{*}{1} & ResNet-50 \cite{he2016deep} & manual & - & 86.4 & 91.6 & 89.3 & 74.6 \\
 & ResNet-18 \cite{he2016deep} & manual & - & 84.4 & 92.1 & 88.9 & 76.2 \\
 & Auto-sklearn \cite{feurer2015efficient} & GO & - & 18.6 & 56.3 & 67.6 & 60.1 \\
 & AutoKeras \cite{jin2019auto} & GO & - & 86.4 & 92.9 & 91.5 & 80.3 \\
 & AutoML & GO & - & 81.1 & 81.8 & 86.1 & 70.6 \\
 & SI-EvoNAS \cite{zhang2020efficient} & EA & - & 90.5±0.7 & 93.0±0.3 & 91.8±0.4 & 80.1±0.3 \\
\hline
\multirow{4}{*}{4} & Single-Tasking (Baseline) \cite{liao2023emt}* & EA & +00.0 & 88.36±1.41 & 93.64±0.41 & 91.18±0.48 & 80.42±0.52 \\
 & EMT-NAS \cite{liao2023emt}* & EMTO & \textbf{-9.55} & 88.60±3.09 & 94.32±0.44 & 91.40±0.73 & 80.72±0.94 \\
 & KTNAS w/o transfer rank & EMTO & +11.27 & 90.04±0.85 & 94.22±0.39 & 91.50±0.48 & 80.57±0.73 \\
 & KTNAS (Ours) & EMTO & -7.93 & \textbf{90.54±1.31} & \textbf{94.38±0.20} & \textbf{91.57±0.38} & \textbf{80.80±0.63} \\
\hline
\end{tabular}
\label{tab:med}
\end{center}
\end{table*}

For test accuracy of discovered  models, KTNAS beats EMT-NAS on Path (90.54\% vs. 88.60\%), Organ\_A (94.38\% vs. 94.32\%), Organ\_C (91.57\% vs. 91.40\%) and Organ\_S (80.80\% vs. 80.72\%).
From ablation study, transfer rank plays a vital role in reducing search cost and improving model performance in multi-task scenarios.

\subsection{Transfer performance analysis}
This section provides deeper insights into transfer performance over 3 multi-task NAS.
We set terminated generation to 100 and run 10 times with different seeds on NAS201.
The validation accuracy and transferred architecture rank on each generation are recorded and the average results of 10 runs are presented.

\begin{figure*}[h]
    \centering
    \begin{subfigure}[b]{0.3\textwidth}
        \includegraphics[width=\textwidth]{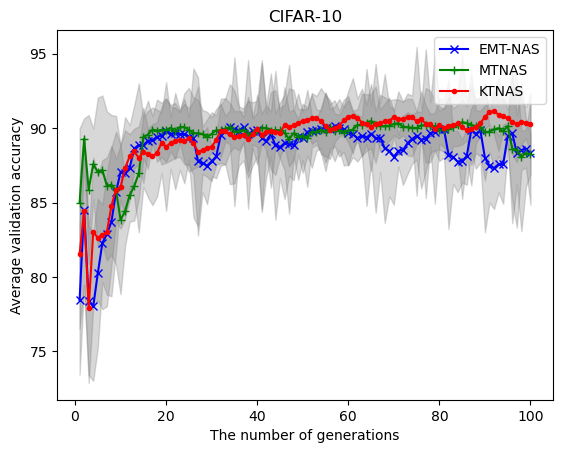}
    \end{subfigure}
    \begin{subfigure}[b]{0.3\textwidth}
        \includegraphics[width=\textwidth]{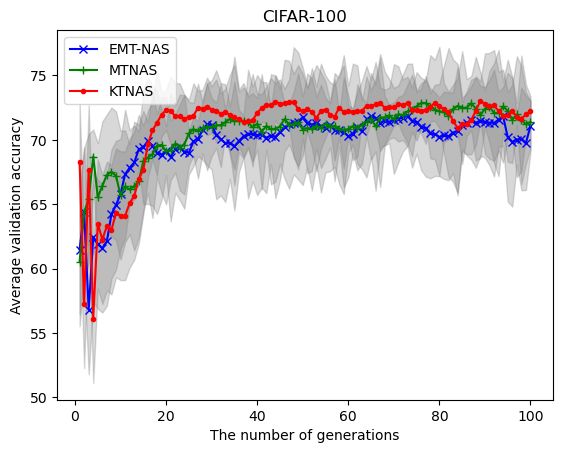}
    \end{subfigure}
    \begin{subfigure}[b]{0.3\textwidth}
        \includegraphics[width=\textwidth]{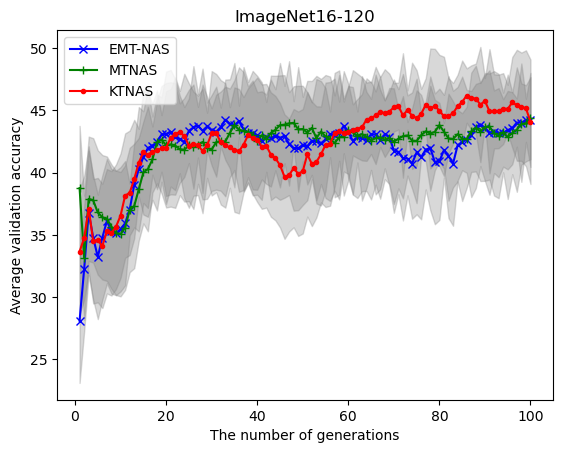}
    \end{subfigure}
    \caption{The average convergence curves on NAS201. The shaded area denotes standard variance over 10 runs.}
    \label{fig:conver_curves}
\end{figure*}

The average convergence curves is shown in Fig. \ref{fig:conver_curves}.
Over different tasks, we observe that parent architectures have significant performance improvement around 20th generation with stable convergence after 60th generation. 
Over distinct algorithms, KTNAS outperforms EMT-NAS and MTNAS on the final and most of intermediate accuracies, demonstrating the effectiveness of our approach in enhancing task performance.
% average convergence curves
% average transferred arch rank
\begin{figure*}[h]
    \centering
    \begin{subfigure}[b]{0.3\textwidth}
        \includegraphics[width=\textwidth]{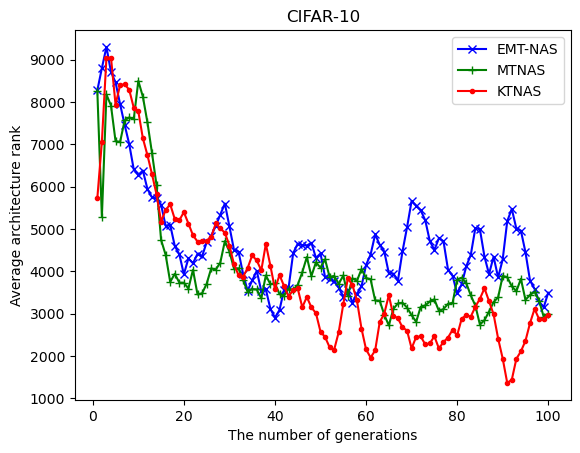}
    \end{subfigure}
    \begin{subfigure}[b]{0.3\textwidth}
        \includegraphics[width=\textwidth]{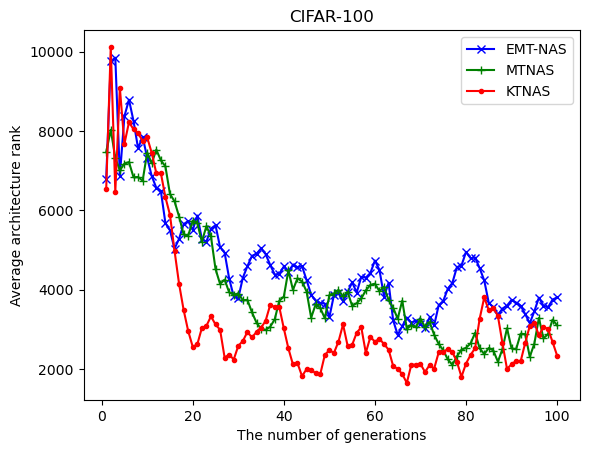}
    \end{subfigure}
    \begin{subfigure}[b]{0.3\textwidth}
        \includegraphics[width=\textwidth]{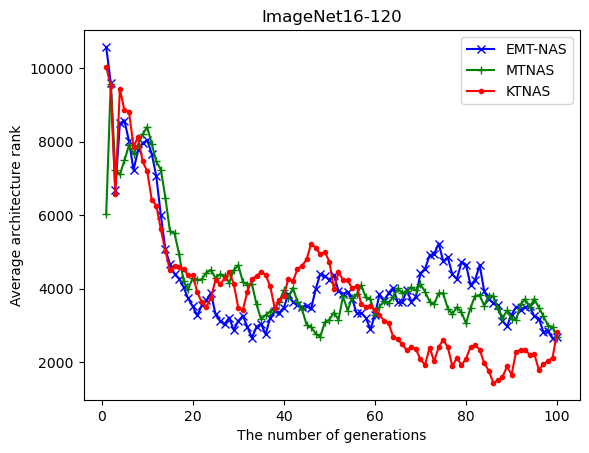}
    \end{subfigure}
    \caption{The average rank of transferred architectures on NAS201.}
    \label{fig:arch_rank}
\end{figure*}

The average transferred architecture rank is shown in Fig. \ref{fig:arch_rank}.
Over different tasks, transferred architectures show an obvious evolution trend in performance rank, indicating knowledge transfer is facilitated by self-evolution processes. 
Over distinct algorithms, KTNAS achieves the lowest average architecture rank compared with peers, claiming the vital role of transfer rank in selecting promising architectures.

\subsection{Parameter sensitivity analysis}\label{param-sense}
Transfer parameter sensitivity results are given in Table \ref{tab:param-sense}. The best results on both validation accuracy and test accuracy are achieved when $r\%=20\%, m=5, M=4$.
% tab:param-sense
\begin{table*}[h]
\captionsetup{justification=centering}
\caption{Comparison of KTNAS when transfer parameters are set to different values on DARTs.}
\begin{center}
\begin{tabular}{|c|c|c|c|c|c|}
\hline
\multirow{2}{*}{Parameter} & \multirow{2}{*}{value} & \multicolumn{2}{|c|}{C-10} & \multicolumn{2}{|c|}{C-100} \\
\cline{3-4}
\cline{5-6}
 &  &  Val Acc(\%) & Test Acc(\%) & Val Acc(\%) & Test Acc(\%) \\ 
\hline
Ranking & 20\% & \textbf{90.84±0.46} & \textbf{97.15±0.04} & \textbf{65.59±0.13} & \textbf{82.88±0.19} \\
ratio & 50\% & 90.83±0.20 & 97.10±0.07 & 65.25±1.32 & 82.54±0.06 \\
 & 80\% & 90.79±0.71 & 96.13±0.05 & 64.84±1.43 & 81.66±0.15 \\
\hline
\# Saved & 1 & 90.01±2.55 & 96.79±0.13 & 65.47±0.86 & 82.32±0.21  \\
generations & 3 & 90.83±0.20 & 97.10±0.07 & 65.25±1.32 & 82.54±0.06  \\
 & 5 & \textbf{90.81±0.86} & \textbf{97.27±0.12} & \textbf{65.52±0.39} & \textbf{83.97±0.19}  \\
\hline
\# Transferred  & 2 & 89.71±1.27 & 96.71±0.06 & 65.06±1.42 & 82.37±0.14 \\
individuals & 4 & \textbf{90.83±0.20} & \textbf{97.10±0.07} & \textbf{65.25±1.32} & \textbf{82.54±0.06} \\
 & 8 & 90.27±0.31 & 96.86±0.03 & 64.98±0.37 & 82.06±0.11 \\
\hline
\end{tabular}
\label{tab:param-sense}
\end{center}
\end{table*}

The choice of graph embedding methods, including those tailored for architecture embedding \cite{yan2020does}\cite{yan2021cate}, is shown in Table \ref{tab:graph}.
We finally choose node2vec as architecture embedding method since the lowest total cost.
% tab:graph
\begin{table}[h]
\captionsetup{justification=centering}
\caption{Comparison on the number of evaluations on NAS201 using different graph embedding methods.}
\label{tab:graph}
\begin{center}
\begin{tabular}{|c|c|c|c|c|}
\hline
Graph Embedding & C-10 & C-100 & ImageNet & Total \\
\hline
arch2vec \cite{yan2020does} & 105.0 & 26.6 & 64.4 & 196.0 \\
CATE \cite{yan2021cate} & 100.5 & 34.1 & 59.2 & 193.8 \\
node2vec \cite{grover2016node2vec} & \textbf{100.7} & \textbf{27.9} & \textbf{60.0} & \textbf{188.5}\\
\hline
\end{tabular}
\end{center}
\end{table}

\section{Conclusion}
In this paper, transfer rank is introduced into multi-task NAS for distinguishing promising architectures over different tasks. 
In addition, we convert architecture into graph and leverage embedding vectors to predict performance of downstream tasks. 
Extensive experiments demonstrate the performance improvement and convergence acceleration of KTNAS compared with other multi-task NAS algorithms. 

% references
\bibliographystyle{IEEEtran}
\bibliography{IEEEabrv,mylib}

\end{document}